%% file: main.tex
\pdfoutput=1
\documentclass[10pt,twocolumn,letterpaper]{article}

\usepackage[pagenumbers]{wacv}

\input{preamble}

\definecolor{linkblue}{rgb}{0.21,0.49,0.74}
\usepackage[pagebackref,breaklinks,colorlinks,allcolors=linkblue]{hyperref}

\title{Tornado4D: Physics-Guided 4D Gaussian Reconstruction of a Laboratory Tornado from Sparse One-Sided Videos}

\author{Yucheng Zhu \qquad Maria J. Molina \qquad Christopher Metzler\\
University of Maryland, College Park}

\begin{document}
\maketitle

\begin{strip}
    \centering
    \includegraphics[width=\textwidth]{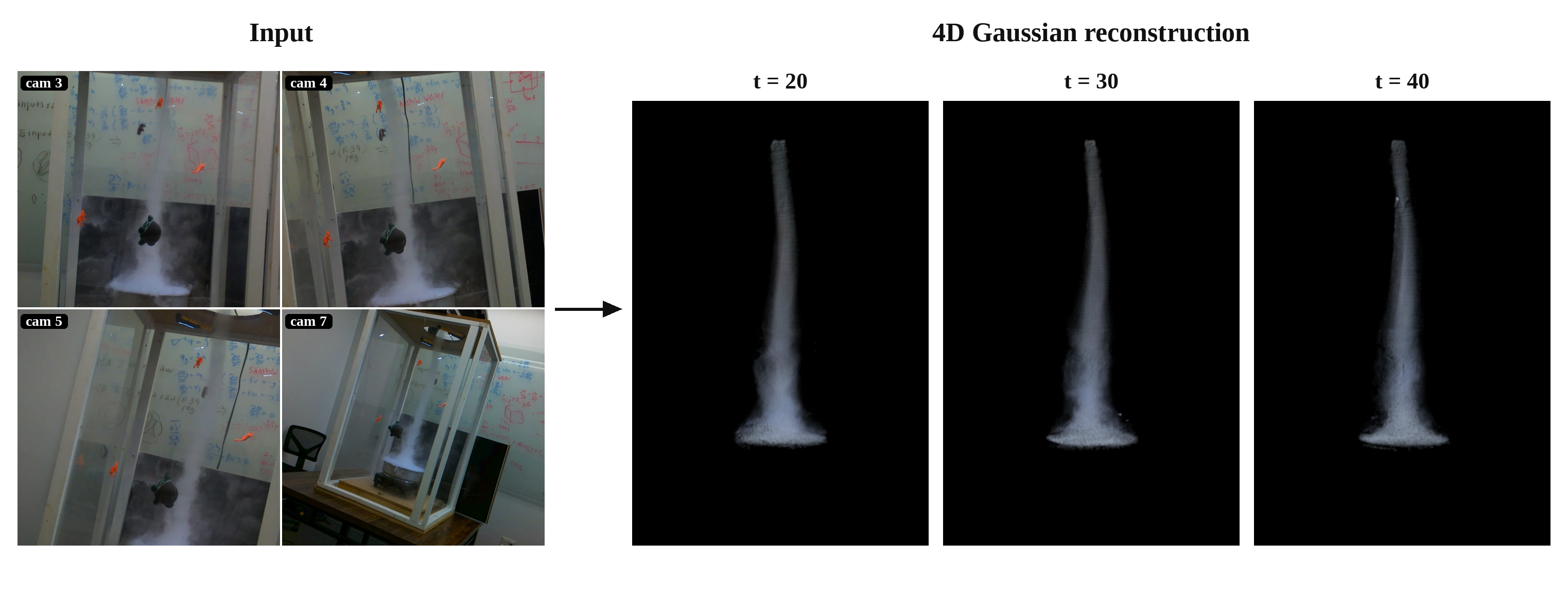}
    \captionof{figure}{\textbf{Tornado4D reconstruction.} The left panel shows four of the input camera views. From these one-sided videos, Tornado4D reconstructs a temporally evolving 3D Gaussian representation, shown at $t=20$, $30$, and $40$ from a fixed novel view.}
    \label{fig:teaser}
\end{strip}

\begin{abstract}
Characterizing how tornadoes evolve over space and time is critically important for understanding and preparing for these highly destructive weather events. Towards this end, we develop algorithms capable of forming time-resolved 3D reconstructions of laboratory smoke tornadoes from sparse one-sided videos. Even in a controlled environment, reconstruction is severely underconstrained: appearance is view dependent, motion is non-rigid, and the unobserved volume admits many geometries. We present Tornado4D, a physics-guided 4D Gaussian framework capable of reconstructing laboratory tornadoes from sparse, one-sided videos. Tornado4D uses knowledge of the enclosure geometry to introduce rotational airflow field priors that constrain how the 3D structure evolves over time. These rotational priors allow us to reconstruct otherwise unobserved structure. We evaluate reconstruction accuracy by computing novel-view synthesis accuracy on a held out camera view: Tornado4D outperforms frame-by-frame 3DGS and video-based 4DGS by a considerable margin.

\end{abstract}

\section{Introduction}
\label{sec:introduction}
Directly observing a tornado's time-varying 3D structure is difficult because tornadoes are short-lived, hazardous, and embedded in larger storms; even rapid-scan mobile Doppler observations require rare favorable deployments \cite{fricker2020evaluating,rotunno2024recent,kosiba2013three}. Laboratory tornado-like vortices provide a controlled alternative in which key behavior can be observed systematically \cite{rotunno2013fluid}. Yang et al.\ introduced a synchronized multi-view dataset of a laboratory smoke tornado and demonstrated static 3DGS reconstruction \cite{yang2025reconstructing}. We use this existing dataset to study a different problem: recovering a time-varying 4D representation of the rotating translucent plume.

Dynamic reconstruction of translucent vortical flow is fundamentally underconstrained. Semi-transparency entangles appearance along camera rays, rapid rotation makes temporal correspondence difficult, and sparse views leave substantial portions of the volume without direct supervision. A flexible model can therefore match the observed images while producing unsupported geometry or physically inconsistent motion.

NeRF and 3DGS support dynamic reconstruction \cite{mildenhall2021nerf,kerbl20233d,wu20244d}, and fluid methods add physical or generative priors \cite{chu2022physics,gao2025fluidnexus,10.1145/3799902.3811188}. Yet independent per-frame reconstruction loses correspondence, generic deformation does not encode the dominant fan-driven transport, and generated views may hallucinate translucent structure. Unlike a freely evolving plume, the laboratory tornado is continuously forced: the top fan drives upward flow, while angled inlet slots induce rotation. We use this known forcing layout to precompute the large-scale airflow rather than asking sparse, one-sided images to infer it from scratch.

To address these ambiguities, Tornado4D couples a precomputed airflow prior derived from this enclosure geometry with differentiable Gaussian reconstruction. The prior supplies coherent base transport where visual evidence is weak, while optimization learns the residual motion needed to explain the observed dynamics. Temporal observations serve two complementary roles: later frames refine the initial state through differentiable advection, and causal sliding windows reconstruct the subsequent motion while preserving particle correspondence.

Our main contributions are threefold:
\begin{itemize}
    \item We refine the underconstrained initial Gaussian state using real later frames by differentiating through precomputed, rotation-dominated advection, without relying on generated views.
    \item We reconstruct the subsequent dynamics by combining base transport from computational fluid dynamics (CFD) with learnable residual motion and causal sliding-window optimization, preserving coherent Gaussian trajectories through rapidly rotating translucent flow.
    \item We evaluate against per-frame 3DGS, 4DGS, and controlled ablations, demonstrating complementary image-space and geometric benefits from canonical refinement and the airflow prior.
\end{itemize}

\section{Related Work}
\label{sec:related-work}

\subsection{Gaussian Splatting for Dynamic and Sparse-View Reconstruction}

3DGS represents scenes with explicit anisotropic Gaussians and differentiable rasterization \cite{kerbl20233d}. Dynamic variants optimize persistent primitive trajectories \cite{luiten2024dynamic}, deform canonical Gaussians at each timestamp \cite{yang2024deformable,wu20244d}, parameterize long-range motion \cite{lin2024gaussian}, or use sliding local canonical models \cite{shaw2024swings}. These flexible models do not explicitly encode fluid transport, so sparse observations may be explained by visually plausible but physically incorrect motion. Yang et al.\ apply standard static 3DGS to a laboratory tornado and introduce the synchronized multi-view dataset used in our experiments \cite{yang2025reconstructing}.

Sparse-view methods use monocular depth guidance \cite{zhu2024fsgs} or diffusion-generated pseudo views \cite{kong2025generative}. Under one-sided capture, generated views may hallucinate unsupported fluid structure. We instead use rotational transport to carry initially hidden Gaussians into regions supervised by real frames.

\subsection{Physics-Guided Fluid Reconstruction}

Physics-informed neural fields constrain density and velocity with fluid equations \cite{chu2022physics}, HyFluid augments a base velocity field with vortex particles \cite{yu2023inferring}, and PAC-NeRF couples a radiance field to Lagrangian particles and MPM simulation \cite{li2023pacnerf}. LPO optimizes PAC-NeRF's initial particles through video to improve first-frame geometry under sparse views \cite{kaneko2024improving}. Our canonical optimization shares this multi-frame principle but targets rotation-dominated smoke represented by 3D Gaussians rather than general continuum materials with NeRF--MPM.

Gaussian fluid methods learn divergence-free velocities, incompressible trajectories, or volume-preserving flows \cite{li2025freegave,du2025gaussfluids,zhang2026gausmoke}. GauSmoke focuses on simple backgrounds and identifies foreground separation in textured scenes as a limitation. FluidNexus combines generated views with physics-integrated particles and dataset-specific diffusion adaptation \cite{gao2025fluidnexus}. Most related, LagrangianSplats transports a canonical Gaussian volume and propagates gradients through sliding windows \cite{10.1145/3799902.3811188}. Our setting additionally provides a known external forcing layout: a fan and angled inlet slots continuously drive the gas. Rather than infer this dominant transport entirely from images, we precompute an enclosure-scale airflow, learn particle residuals for local mismatch, and exploit rotation for real-view supervision of initially hidden geometry.

\subsection{Video Matting for Transparent Foreground Extraction}

Video matting decomposes frames into foreground, background, and continuous alpha, preserving transparency discarded by binary masks. RVM and MatAnyone provide temporally guided propagation \cite{lin2022robust,yang2025matanyone}; VideoMaMa refines coarse masks into detailed, consistent mattes with a pretrained video prior \cite{Lim_2026_CVPR}. We use VideoMaMa to preprocess the translucent tornado.

\section{Method}
\label{sec:method}

\begin{figure*}[t]
    \centering
    \includegraphics[width=\textwidth]{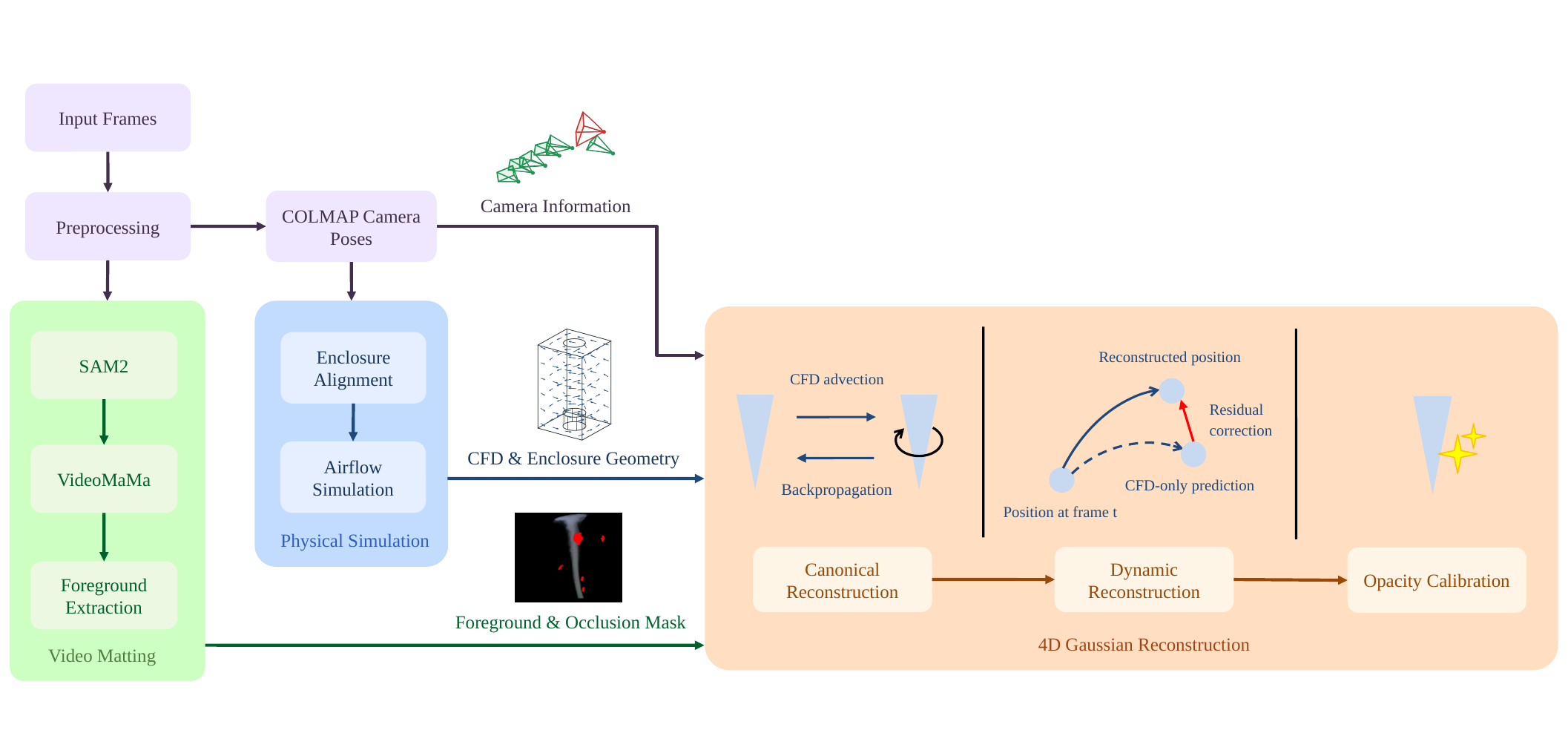}
    \caption{\textbf{Tornado4D pipeline.} Matting provides a translucent tornado layer, and enclosure geometry is used to simulate the airflow field. Advection-guided multi-frame supervision refines the canonical state, causal dynamic reconstruction combines CFD advection with learned residuals, and post-training calibration adjusts the opacity of the reconstructed tornado.}
    \label{fig:method-overview}
\end{figure*}

\subsection{Problem Formulation}

Let $\mathcal{V}=\{1,\ldots,7\}$ index the cameras, $\mathcal{V}_{\mathrm{train}}\subset\mathcal{V}$ the views available to reconstruction, $I_{v,t}$ their observations, and $\pi_v$ their known parameters. Remaining views are held out for evaluation. Alpha matting and a clean background produce the black-background target $Y_{v,t}$. The enclosure geometry defines $\Omega\subset\mathbb{R}^3$ and a fixed qualitative airflow prior $u_{\mathrm{CFD}}:\Omega\rightarrow\mathbb{R}^3$.

The output is a sequence $\mathcal{G}_{0:T}$ of 3D Gaussian sets. At time $t$,
\begin{equation}
    \mathcal{G}_t=\{g_{i,t}\}_{i=1}^{N_t}, \qquad
    g_{i,t}=(x_{i,t},s_{i,t},q_{i,t},o_{i,t},c_{i,t}),
    \label{eq:dynamic-gaussians}
\end{equation}
where $x$, $s$, $q$, $o$, and $c$ denote position, scale, orientation, opacity, and view-independent RGB color. Here $N_t$ is the effective active population within a fixed capacity; birth factors and lifecycle opacities determine which particles contribute. With differentiable renderer $\mathcal{R}$, we seek
\begin{equation}
    \mathcal{R}(\mathcal{G}_t;\pi_v)\approx Y_{v,t}
    \quad \text{for } v\in\mathcal{V}_{\mathrm{train}},\ t=0,\ldots,T,
    \label{eq:rendering-objective}
\end{equation}
while anchoring motion to $u_{\mathrm{CFD}}$ through particle-wise velocity residuals. The aim is temporally coherent, enclosure-consistent geometry and motion despite one-sided supervision.

\subsection{System Overview}

As shown in \cref{fig:method-overview}, Stage I initializes $\mathcal{G}_0$ with static Gaussian splatting and advects it through nearby frames. Multi-frame losses update canonical geometry as hidden Gaussians rotate into observed regions. Stage II expands this state and reconstructs the sequence with causal sliding windows, fixed CFD transport, and learned residual velocities. Stage III freezes geometry and motion and calibrates the opacity of the Gaussian population inherited from Stage I.

\subsection{Foreground Processing and Airflow Prior}
\label{sec:foreground-flow}

\paragraph{Foreground target construction.} We stabilize and color-correct the videos before COLMAP calibration. For each camera video, we manually annotate point prompts for the tornado and foreground occluders in one frame, and SAM 2 \cite{ravi2025sam} propagates both masks throughout the video. VideoMaMa then converts the tornado masks into alpha mattes. Given composite frame $I_{v,t}$, alpha matte $A_{v,t}$, and clean background $B_v$, we recover the foreground and form the black-background reconstruction target as
\begin{equation}
    \begin{aligned}
        F_{v,t}
        &= \mathrm{clip}_{[0,1]}\!\left(
            \frac{I_{v,t}-(1-A_{v,t})\odot B_v}
                 {\max(A_{v,t},\epsilon)}
          \right), \\
        Y_{v,t} &= A_{v,t}\odot F_{v,t}.
    \end{aligned}
    \label{eq:foreground-recovery}
\end{equation}

Let $M_{v,t}$ be zero inside the union of occluder masks and one elsewhere. We mask both $L_1$ and SSIM. Invalid rendered pixels are replaced by detached target values before evaluating SSIM, preventing gradients from entering masked render pixels. Occluded pixels therefore contribute no direct reconstruction gradients.

\begin{figure}[t]
    \centering
    \includegraphics[width=\columnwidth]{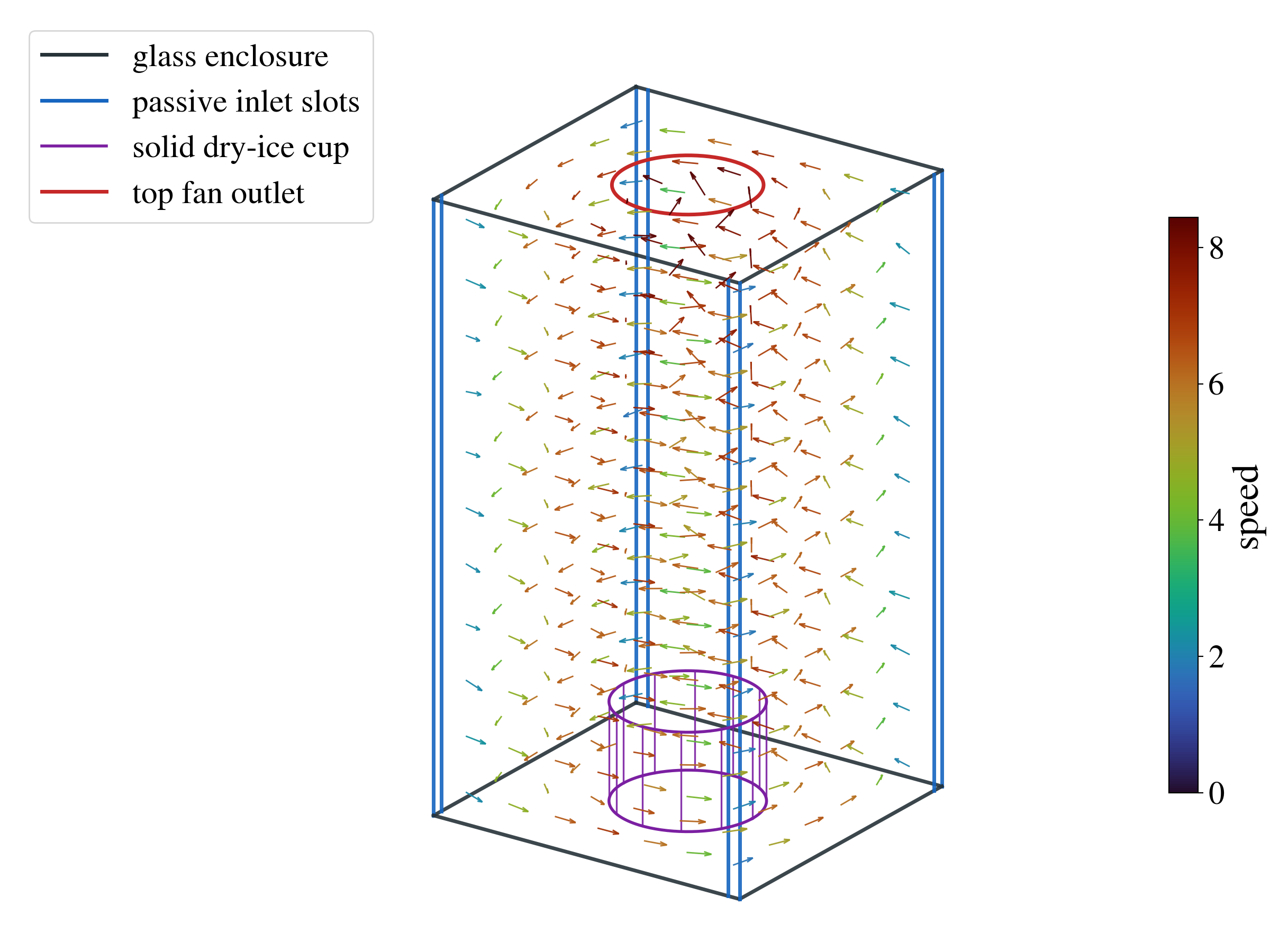}
    \caption{\textbf{Geometry-derived airflow prior.} Arrows show simulated velocity direction in local enclosure coordinates, and color indicates speed. Lines identify the rectangular boundary, four passive wall slots, solid dry-ice cup, and top-fan outlet. The field captures the dominant rotation and upward transport used by both reconstruction stages, while learned residual velocities correct local mismatch to the observed smoke.}
    \label{fig:airflow-prior}
\end{figure}

\paragraph{Geometry-derived airflow prior.} Figure~\ref{fig:airflow-prior} visualizes the fixed three-dimensional velocity field used by our reconstruction. We import a preliminary 3DGS reconstruction into 3D modeling software and manually align a rectangular simulation volume with the glass enclosure, its four vertical inlet slots, top fan, and dry-ice cup. This aligned geometry defines the simulation domain and maps the simulated field to the camera coordinate system.

In the same local coordinate frame, we initialize the velocity to zero and solve the incompressible flow model
\begin{equation}
    \frac{\partial u}{\partial \tau}
    +(u\cdot\nabla)u
    =-\nabla p+\nu\nabla^2u-\lambda u,
    \qquad \nabla\cdot u=0,
    \label{eq:airflow-model}
\end{equation}
We discretize this model on a staggered marker-and-cell (MAC) grid~\cite{harlow1965numerical}. Following the semi-Lagrangian projection framework of Stam~\cite{stam2023stable}, we advect velocity by backtracing, apply explicit viscosity and linear drag, and project the result to enforce discrete incompressibility while preserving the prescribed boundary fluxes. We use monotone BFECC error compensation to reduce numerical dissipation in velocity advection~\cite{liu2005flowfixer}. The top fan is the sole active driver and extracts air through a circular aperture. Four full-height wall slots provide flux-matched inflow whose tangential component induces rotation; the other glass faces and the cup geometry are impermeable. We omit heating, buoyancy, smoke density, and two-way coupling. The simulation uses prescribed inlet geometry, viscosity, and linear drag, while the mean outlet-normal velocity controls the overall boundary-flow magnitude.

Starting from zero velocity, we iterate the solver to convergence. The resulting field is trilinearly sampled, transformed to COLMAP coordinates, and frozen in both reconstruction stages. It is a qualitative transport prior; image supervision and residual velocity absorb local errors.

\subsection{Advection-Guided Canonical Reconstruction}
\label{sec:canonical-reconstruction}

\paragraph{Static initialization.} We take the first reconstruction frame as $t=0$ and uniformly initialize isotropic Gaussians in a cylinder spanning the enclosure height, with radius equal to half its smaller horizontal dimension. A static Gaussian-splatting optimization learns their position, radius, opacity, and color from the available training views, with adaptive densification and opacity pruning. Throughout both reconstruction stages, our image objective is
\begin{equation}
    \mathcal{L}_{\mathrm{img}}
    = \lambda_1\mathcal{L}_1
    +\lambda_{\mathrm{SSIM}}(1-\operatorname{SSIM}),
    \label{eq:image-loss}
\end{equation}
evaluated only over valid pixels as described in \cref{sec:foreground-flow}. This static reconstruction provides plausible observed-view appearance but leaves the side and rear geometry underconstrained.
The static objective additionally penalizes mean-squared distance outside the enclosure volume.

\begin{figure}[H]
    \centering
    \includegraphics[width=\linewidth]{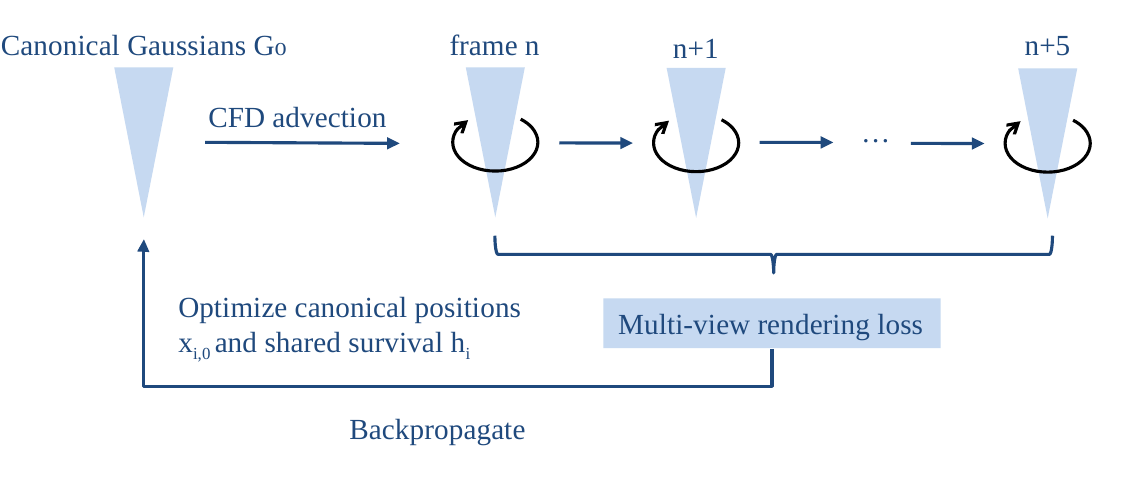}
    \caption{\textbf{Advection-guided canonical reconstruction.} Multi-view losses backpropagate through CFD advection to frame-0 positions and survival variables. Rotation exposes initially hidden Gaussians to real observations.}
    \label{fig:canonical-refinement}
\end{figure}

\paragraph{Advection-guided canonical refinement.} Let $\Phi_{0\rightarrow t}$ denote differentiable advection from frame 0 to $t$ under the fixed field $u_{\mathrm{CFD}}$. We advect the canonical Gaussians over a short temporal prefix and supervise overlapping temporal windows $\mathcal{W}_k$. Only the canonical positions $\{x_{i,0}\}$ and a time-shared survival logit $h_i$ are optimized; all other Gaussian attributes remain frozen. The effective opacity of canonical Gaussian $i$ is $o_i\sigma(h_i)$. Writing $\ell_{v,t}(\mathcal{G})= \mathcal{L}_{\mathrm{img}}(\mathcal{R}(\mathcal{G};\pi_v),Y_{v,t})$, the objective for window $k$ is
\begin{equation}
    \begin{aligned}
    \mathcal{L}_{\mathrm{can}}^{(k)}
    &= \frac{1}{|\mathcal{W}_k|\,|\mathcal{V}_{\mathrm{train}}|}
       \sum_{t\in\mathcal{W}_k}\sum_{v\in\mathcal{V}_{\mathrm{train}}}
       \ell_{v,t}\!\left(\Phi_{0\rightarrow t}(\mathcal{G}_0)\right) \\
    &\quad +\lambda_h\operatorname{mean}_i\bigl(1-\sigma(h_i)\bigr).
    \end{aligned}
    \label{eq:canonical-objective}
\end{equation}
\Cref{fig:canonical-refinement} illustrates this optimization.
Each sweep visits the windows sequentially and takes one optimizer step per window. The retention term discourages unnecessary removal while allowing unsupported Gaussians to be suppressed. More importantly, rotation carries Gaussians from the unobserved side into camera-facing regions, so later-frame gradients update their frame-0 positions without generated views.

\subsection{Physics-Guided Dynamic Particle Reconstruction}
\label{sec:dynamic-reconstruction}

\begin{figure}[H]
    \centering
    \includegraphics[width=0.7\linewidth]{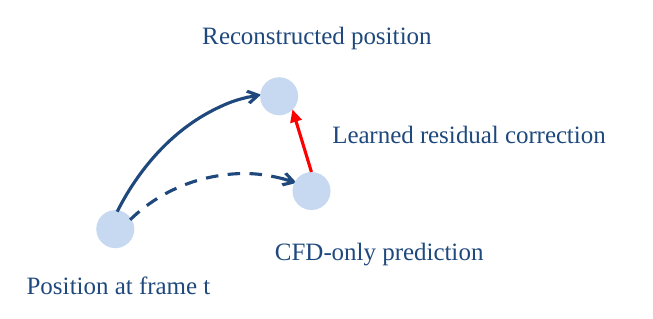}
    \caption{\textbf{Physics-guided residual motion.} The spheres indicate the positions of a single Gaussian. A learned velocity residual corrects CFD-predicted displacement under visual supervision.}
    \label{fig:residual-dynamics}
\end{figure}

\paragraph{Residual motion model.} Starting from the optimized $\mathcal{G}_0$, each particle and frame transition has a learnable velocity residual $\Delta v_{i,t}$:
\begin{equation}
    v_{i,t}(x)=u_{\mathrm{CFD}}(x)+\Delta v_{i,t}.
    \label{eq:residual-velocity}
\end{equation}
We integrate this field with an RK2 midpoint step,
\begin{equation}
    x^{\mathrm{mid}}_{i,t}
    =x_{i,t}+\frac{\Delta t}{2}v_{i,t}(x_{i,t}),\qquad
    x_{i,t+1}
    =x_{i,t}+\Delta t\,v_{i,t}(x^{\mathrm{mid}}_{i,t}),
    \label{eq:rk2-advection}
\end{equation}
using adaptive RK2 substeps that limit the maximum displacement relative to the CFD grid spacing within each video frame. Thus, the CFD field provides large-scale rotation and upward transport, while visual gradients use $\Delta v_{i,t}$ to correct local simulation error, as illustrated in \cref{fig:residual-dynamics}.

\paragraph{Causal sliding-window optimization.} We refer to the Gaussians obtained from the canonical reconstruction as the initial Gaussians. We expand this set by sampling parents in proportion to their effective frame-0 opacity, applying support-local jitter, and correcting opacity to preserve compositing mass. We also allocate a fixed cohort of source candidates to each future frame while maintaining a fixed total particle capacity. We optimize successive fixed-length windows; after each window converges, its oldest transition is committed and the resulting state is detached before advancing the window, yielding fixed-lag truncated backpropagation. The first window additionally optimizes the positions of the initial Gaussians; subsequent windows optimize the uncommitted velocity residuals and per-frame birth-opacity factors.

In addition to \cref{eq:image-loss}, we penalize temporal variation and differences between neighboring particles:
\begin{equation}
    \mathcal{L}_{\mathrm{dyn}}
        = \mathcal{L}_{\mathrm{img}}
        +\lambda_t\mathcal{L}_{\mathrm{temp}}
        +\lambda_s\mathcal{L}_{\mathrm{spatial}}.
    \label{eq:dynamic-objective}
\end{equation}
Here $\mathcal{L}_{\mathrm{temp}}$ and $\mathcal{L}_{\mathrm{spatial}}$ penalize residual differences across adjacent transitions and canonical-neighborhood edges. These terms preserve local flexibility while the frozen CFD field supplies large-scale transport. Frame-0 positions remain image-supervised in the first window. A trajectory leaving the enclosure outside the fan aperture immediately receives zero opacity.

\paragraph{Particle birth and lifecycle.} For each future frame, we place a cohort of source candidates on a thin disk above the cup. Each frame's candidates share a learnable birth opacity factor and remain exactly transparent before birth, preventing future observations from activating smoke prematurely. Thereafter they follow the same motion model and persist until reaching the fan or another enclosure boundary, at which point they become permanently transparent.

\paragraph{Cup-aware rendering and trajectory filtering.} An analytic cylindrical depth occluder prevents Gaussians behind the cup from contributing to rendered pixels, and canonical updates entering the solid are rejected. Separately, after optimization we constrain exported trajectories to a prescribed cylindrical support above the cup. A trajectory leaving this support is suppressed in every frame, and the resulting time-shared mask is applied consistently during rendering and export.

\subsection{Post-Training Opacity Calibration}
\label{sec:opacity-calibration}

The initial Gaussian opacities are inherited from the static reconstruction and remain fixed during the preceding stages. Because the learned survival factor lies in $[0,1]$, these stages can only reduce each initial Gaussian's effective opacity and cannot increase it, causing the reconstructed tornado to become visibly darker. After all causal windows are committed, we therefore freeze geometry and motion and calibrate the brightness of the initial Gaussians with a single Beer--Lambert density gain $g$. Let $\alpha_i=o_i\sigma(h_i)$ be the effective opacity of an initial Gaussian $i$. We apply
\begin{equation}
    \alpha_i^{\prime}=1-(1-\alpha_i)^g.
    \label{eq:opacity-calibration}
\end{equation}
This scales optical density and preserves compositing behavior. With $h_i$ frozen, we convert $\alpha_i^{\prime}$ back to a compatible base opacity and bake it into the initial Gaussian state; the opacities of Gaussians emitted after initialization remain unchanged. We freeze all other variables and choose $g$ by bounded deterministic log-domain search over the held-in image objective with weak log-gain regularization. A coarse grid followed by golden-section refinement determines the gain.

\section{Experiments}
\label{sec:experiments}

\subsection{Data Split and Evaluation Protocol}

\begin{figure}[H]
    \centering
    \includegraphics[width=\linewidth]{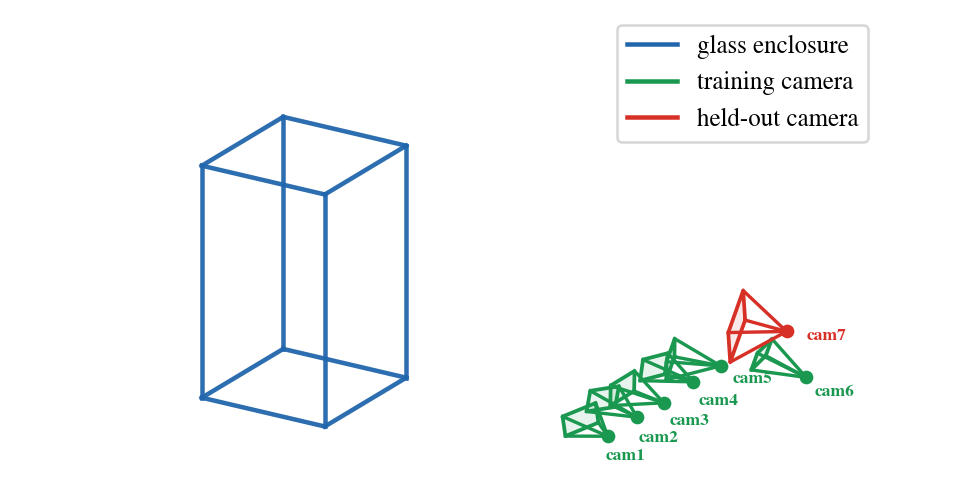}
    \caption{\textbf{Camera layout.} Orthographic view of the camera poses and glass enclosure. Held-out camera 7 (red) lies beyond the six-camera training arc.}
    \label{fig:camera-layout}
\end{figure}

We reconstruct $t=0$--$40$ at 30 FPS from the sequence of Yang et al.\ \cite{yang2025reconstructing}. Of the eight capture cameras, seven were successfully registered by COLMAP and are used in our experiments. These cameras span a one-sided $39.2^\circ$ arc. We reserve endpoint camera 7 for held-out evaluation; it lies $7.46^\circ$ beyond its nearest training camera (\cref{fig:camera-layout}). For quantitative evaluation, all methods are trained on the remaining six cameras; the spatial and temporal qualitative comparisons use separate models trained with all cameras.

We evaluate uncropped $1920\times1080$ camera 7 foregrounds, excluding pixels identified by the SAM 2 occluder masks from PSNR, SSIM, and AlexNet LPIPS. All methods share observations, masks, the six-camera training split, frame range, and evaluation protocol; our variants also share trajectory filtering and opacity calibration.

\subsection{Implementation Details}

The airflow solver uses an $80\times144\times80$ grid, pseudo-time step $1.25\times10^{-2}$, viscosity $4\times10^{-3}$, linear drag $2\times10^{-2}$, and inlet angle $70^\circ$. We set the outlet-normal velocity to $8.0$ COLMAP units/s. With a 4,000-step limit, the convergence criteria are met after 3,000 iterations.

Stage I initializes 50k Gaussians and runs static reconstruction for 1,500 iterations. We set $(\lambda_1,\lambda_{\mathrm{SSIM}})=(8\times10^{-1},2\times10^{-1})$ and the enclosure-boundary weight to $1\times10^{-1}$. Canonical refinement covers $t=0$--$10$ with six-frame windows and runs for 100 sweeps. The position and survival learning rates are $1\times10^{-3}$ and $5\times10^{-2}$, respectively, with $\lambda_h=1\times10^{-3}$. Particle upsampling uses jitter fraction $0.75$. We allocate 350 source candidates to each of 40 future frames, with initial birth factor $0.65$ and maximum opacity $0.1$, while fixing total particle capacity at 100k.

Stage II spans 41 frames using 37 five-frame windows and optimizes each window for 30 epochs. The residual, initial-position, and birth-factor learning rates are $3\times10^{-2}$, $2\times10^{-4}$, and $1\times10^{-2}$, respectively. We use $(\lambda_t,\lambda_s)=(1\times10^{-2},3\times10^{-3})$, eight spatial neighbors per particle, and an outlet margin of $0.1$. In enclosure-aligned coordinates, the trajectory support uses radius $1.139$ and lower vertical bound $-2.98$.

Stage III performs post-training opacity calibration. We search $g\in[0.25,4]$ using a nine-point grid and eight golden-section refinement steps, with log-gain regularization $1\times10^{-3}$.

For 3DGS \cite{kerbl20233d}, we independently reconstruct each frame for 30k iterations using its official initialization and retain its gradient-based densification schedule from iteration 500 through 15k. 4DGS \cite{wu20244d} uses its official multi-view initialization and training schedule of 3,000 coarse iterations followed by 15k fine iterations, with its original gradient-based densification and pruning. Both baselines use the same six training cameras and hold out camera 7.

All experiments are run on a single NVIDIA RTX A6000 GPU. Precomputing the airflow prior takes approximately 1.5 minutes, and training one 41-frame Tornado4D model through Stages I--III takes approximately two hours. These timings exclude data preprocessing, evaluation rendering, and metric computation.

\subsection{Comparison and Ablations}

We compare Ours with per-frame 3DGS, 4DGS, and two ablations. \emph{Without canonical refinement} sends the static checkpoint directly to Stage II while retaining CFD. \emph{Without canonical refinement and CFD} additionally zeros the base velocity and learns transition velocities under the same smoothness losses. Comparing these variants isolates the effect of CFD when canonical refinement is absent. Both retain particle birth, lifecycle, cup handling, and calibration.

\begin{table}[H]
    \centering
    \caption{\textbf{Held-out test-view reconstruction.} Metrics cover 41 full-resolution foreground frames from camera 7, with pixels belonging to occluders excluded.}
    \label{tab:heldout-comparison}
    \setlength{\tabcolsep}{4.5pt}
    \begin{tabular}{lccc}
        \toprule
        Method & PSNR $\uparrow$ & SSIM $\uparrow$ & LPIPS $\downarrow$ \\
        \midrule
        3DGS \cite{kerbl20233d} & 29.72 & 0.970 & 0.040 \\
        4DGS \cite{wu20244d} & 30.22 & 0.969 & 0.042 \\
        w/o canonical refinement & 34.71 & 0.983 & 0.029 \\
        w/o canonical and CFD & 34.16 & 0.979 & 0.029 \\
        Ours & \textbf{35.66} & \textbf{0.984} & \textbf{0.020} \\
        \bottomrule
    \end{tabular}
\end{table}

\paragraph{Comparison to Gaussian baselines.} Ours leads 3DGS and 4DGS on all three aggregate held-out metrics, demonstrating better reconstruction on the test camera.

\paragraph{Ablation analysis.} Both ablated variants worsen all metrics relative to the full model. Canonical refinement improves aggregate PSNR by $0.94$ dB and matters most near the sequence start, where causal reconstruction has little context. Under the shared setting without canonical refinement, retaining CFD improves PSNR by $0.55$ dB and SSIM from $0.979$ to $0.983$, while their LPIPS values are tied at $0.029$ after rounding. This comparison supports the contribution of coherent transport to image-space reconstruction. Applying the same calibration protocol to all variants limits brightness mismatch as a confounding factor.

\subsection{Novel View Synthesis}

\begin{figure}[t]
    \centering
    \includegraphics[width=\linewidth]{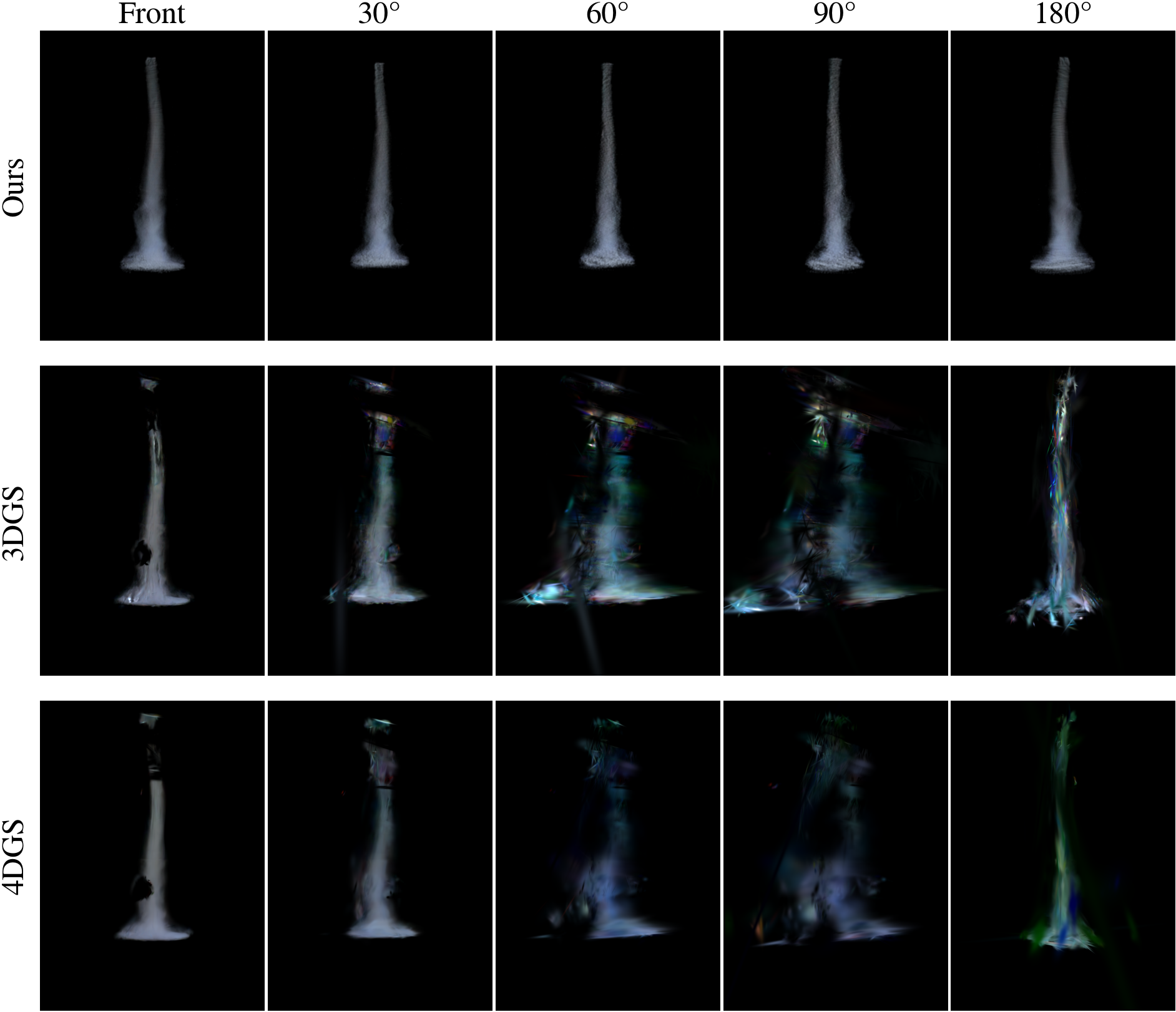}
    \caption{\textbf{Novel-view comparison across methods.} All methods are trained with all cameras and rendered at $t=20$. Columns rotate from the enclosure-facing front view to $180^\circ$. Ours maintains a coherent volumetric funnel across viewpoints, whereas 3DGS and 4DGS exhibit substantial artifacts away from the front.}
    \label{fig:spatial-views}
\end{figure}

We further compare novel-view synthesis using models trained with all cameras. The front view is aligned with the face of the enclosure toward the capture cameras, and the remaining views rotate around the enclosure-aligned vertical axis. As shown in \cref{fig:spatial-views}, Ours preserves a connected funnel, stable base, and consistent opacity from the front through $180^\circ$. In contrast, 3DGS produces prominent floating and stretched structures as the viewpoint rotates, while 4DGS becomes dark and loses coherent plume structure away from the front. Because these rotated viewpoints have no ground-truth images, this comparison evaluates qualitative multi-view consistency rather than image-space accuracy.

\begin{figure}[t]
    \centering
    \includegraphics[width=\linewidth]{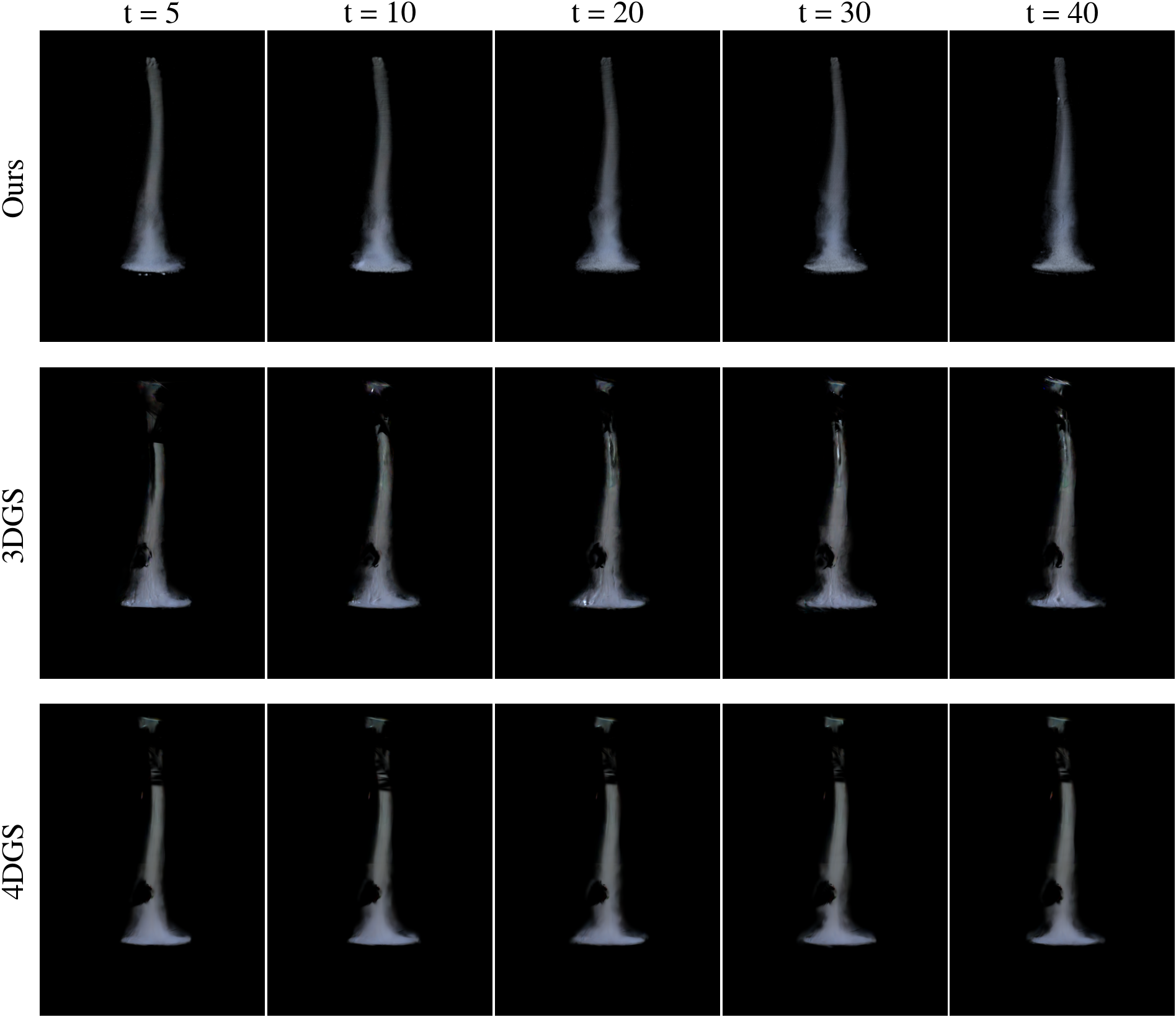}
    \caption{\textbf{Temporal novel-view comparison.} From the fixed enclosure-facing front view, we render Ours, independently trained per-frame 3DGS, and 4DGS at five time points. Ours preserves a coherent funnel throughout the sequence with fewer floating and dark artifacts.}
    \label{fig:temporal-comparison}
\end{figure}

Figure~\ref{fig:temporal-comparison} fixes the same enclosure-facing front viewpoint and renders five times across the reconstructed sequence. Ours maintains a connected tornado while its contour and density evolve over time. The baseline reconstructions are less complete and contain black structures that occlude parts of the tornado. In regions hidden by occluders in the input views, the baselines recover only dark silhouettes, whereas our rotational airflow prior and cross-frame supervision complete the previously unobserved tornado structure. Together with the spatial rotation in \cref{fig:spatial-views}, this result indicates that the physics-guided representation is more stable across both viewpoint and time.

\section{Limitations}
\label{sec:limitations}

We reconstruct and evaluate one controlled laboratory smoke-tornado sequence without time-resolved 3D ground truth. How Tornado4D transfers to real-world tornado observations remains an open question: field data may lack synchronized calibrated views and known boundaries, while operating at vastly different scales and atmospheric conditions. All cameras in our laboratory sequence remain on one side, so camera 7 metrics cannot validate the unseen rear geometry, and the cylindrical-support test does not establish comprehensive physical consistency. Surrounding cameras or volumetric measurements would enable stronger validation.

Our airflow prior is derived from the enclosure geometry but remains qualitative, precomputed, and frozen; image losses cannot refine its boundary conditions. It omits buoyancy, thermal variation, and smoke--air coupling, so inaccurate transport can still bias trajectories despite learned residuals. A differentiable flow model could jointly refine the prior. Results also depend on video matting and fixed appearance, which cannot capture every density or illumination change.

\section{Conclusion}
\label{sec:conclusion}

We introduced Tornado4D to reconstruct a laboratory smoke tornado as a temporally evolving 3D Gaussian representation from sparse, synchronized, one-sided videos. The method combines a geometry-derived airflow prior, differentiable advection for canonical refinement, and causal sliding-window optimization with learned residual motion. Our experiments show that these components recover more accurate held-out views and more consistent particle trajectories than Gaussian baselines and controlled ablations, without relying on generated views. This reconstruction problem is difficult because semi-transparency entangles appearance along camera rays, rapid non-rigid rotation complicates correspondence, and one-sided observations leave much of the volume unconstrained. Addressing these ambiguities is important because observation-derived, time-resolved 3D representations can complement sparse measurements of vortical phenomena and provide a controlled step toward recovering tornado structure from real-world observations.

{
    \small
    \bibliographystyle{ieeenat_fullname}
    \bibliography{main}
}

\end{document}

%% file: preamble.tex
\usepackage{amsmath,amssymb}
\usepackage{booktabs}
\usepackage{graphicx}
\usepackage{float}
\usepackage{subcaption}
\usepackage{cuted}
\usepackage{microtype}